\documentclass{bmvc2k}

\title{Reasmory: 3D Reconstruction as Explicit Memory for VLMs Spatial Reasoning}

\addauthor{Jixuan He}{jh2926@cornell.edu}{1}
\addauthor{Xueting Li}{xuetingli1123@gmail.com}{2}
\addauthor{Chieh Hubert Lin}{hubert052702@gmail.com}{3}
\addauthor{Ming-Hsuan Yang}{minghsuanyang@gmail.com}{4}

\addinstitution{
 Cornell Tech, Cornell University
}
\addinstitution{
 NVIDIA
}
\addinstitution{
illoca AI
}
\addinstitution{
The University of California, Merced
}

\runninghead{HE, Li, Lin and Yang}{Reasmory}

\def\eg{\emph{e.g}\bmvaOneDot}

\usepackage{booktabs}
\usepackage[table]{xcolor}
\usepackage{algorithm}
\usepackage{algpseudocode}
\usepackage{makecell}
\usepackage{multirow}
\begin{document}

\maketitle

\begin{abstract}
Vision-Language Models (VLMs) exhibit emerging spatial reasoning capabilities, yet they remain unreliable on tasks requiring precise spatial understanding, such as viewpoint reasoning, directional comparison, and distance estimation. In multi-view images and monocular videos, relevant spatial cues are often sparse and distributed across redundant observations, making them difficult to organize and exploit. Reconstruction-based Vision Foundation Models (VFMs) offer a natural way to aggregate such observations into explicit spatial memory, such as point clouds. However, simply exposing reconstruction models as free-form tools is brittle, VLMs may invoke tools incorrectly, skip required spatial transformations, or misuse intermediate results.
We propose \textbf{Reasmory}, a framework that formulates spatial reasoning as structured program execution over reconstructed spatial memory. Reasmory constructs explicit 3D memory, augments it with semantically grounded 3D object instances, and introduces a lightweight Domain-Specific Language (DSL) that constrains how VLMs query objects and cameras, transform viewpoints, and render observations during reasoning. Generated programs are parsed and validated before execution, enabling more reliable interaction with spatial memory than unconstrained tool use. Experiments on multi-view image and video spatial reasoning benchmarks show consistent gains of 6--18\% over strong baselines, including GPT-5-mini and Gemini-3-flash, indicating that explicit 3D memory is most useful when accessed through constrained, validated operations rather than free-form tool calls.
\end{abstract}
\section{Introduction}
\begin{figure}[t]
  \centering
   \includegraphics[width=.9\linewidth]
   {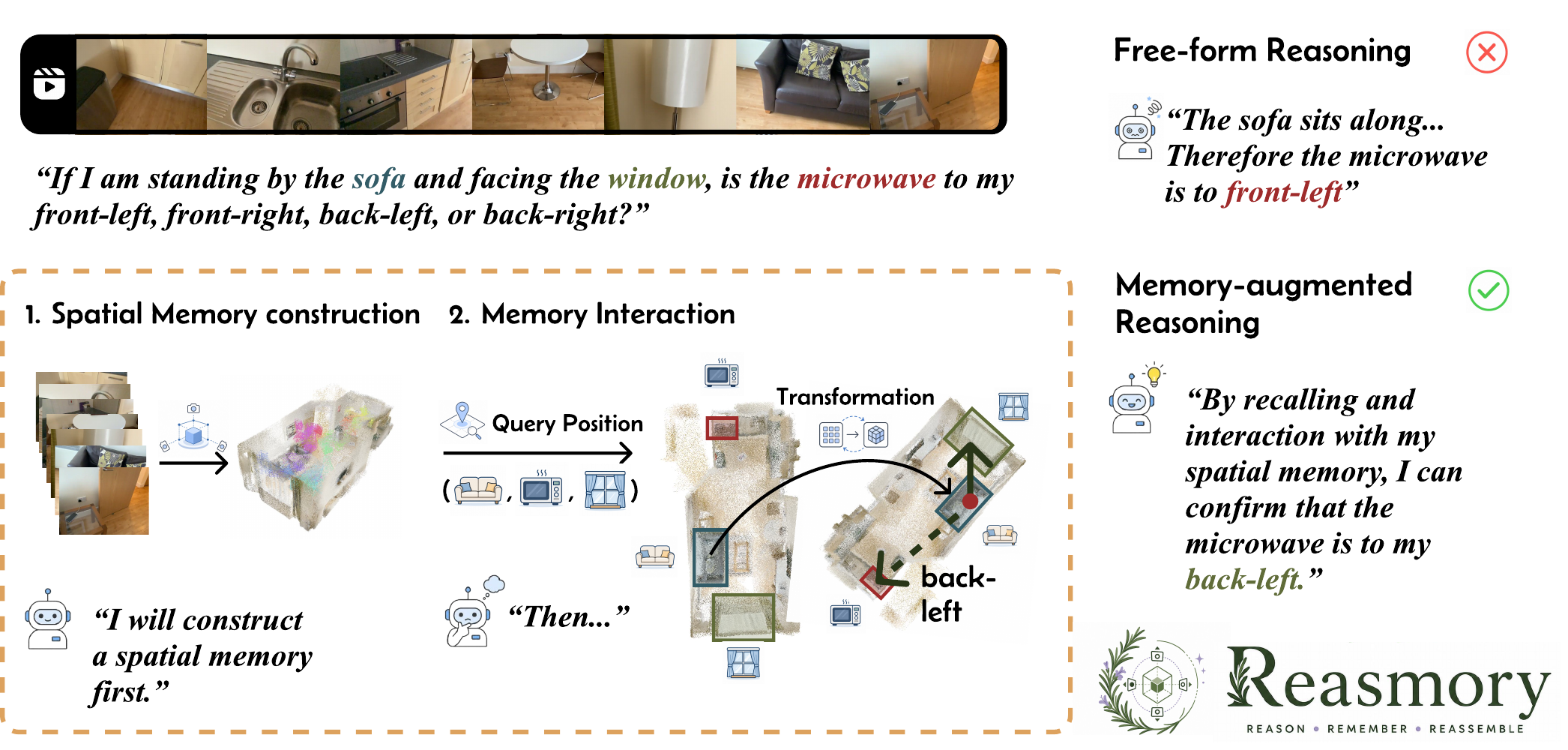} 
   \vspace{-2mm}
   \caption{Overview of \textbf{Reasmory}. Spatial evidence in multi-view images and videos is often sparse and redundant, making it important to organize evidence explicitly for VLM spatial reasoning. Reasmory addresses this by constructing explicit 3D spatial memory and constraining VLM interaction with this memory through validated DSL programs.}
   \label{fig:teaser}
   \vspace{-5mm}
\end{figure}

Understanding spatial relationships is a fundamental capability for intelligent agents. For Vision-Language Models (VLMs), this capability is particularly important in embodied settings, where tasks such as vision-language action~\cite{zitkovich2023rt,kim2024openvla} and navigation~\cite{qi2025vln} require reliable spatial reasoning. Despite recent advances in visual understanding~\cite{liu2023visual,liu2024improved,bai2025qwen3}, VLMs remain unreliable when reasoning across multi-view images and videos. As shown in Figure~\ref{fig:teaser}, these tasks require perceiving relevant objects, retaining observations across time or viewpoints, constructing coherent spatial representations, and performing spatial transformations such as viewpoint changes, directional comparisons, or distance estimation. This perception--recall--reasoning pipeline is inherently fragile: failures in perception or memory can corrupt the spatial representation, while errors in spatial transformation can produce incorrect answers even when sufficient evidence is available.

We show that reliable spatial reasoning depends on how spatial information is organized, retrieved, and used during reasoning. This challenge is especially evident in multi-view image and video settings, where relevant spatial cues are sparse and scattered across redundant observations. As a result, adding more frames can increase visual clutter and hinder recall rather than improve spatial understanding. To study this effect, we conduct controlled experiments on 200 samples from VSI-Bench~\cite{yang2025thinking} using SpaceOM~\cite{VQASynth} under a fixed budget of 16 frames. Question-conditioned CLIP-based frame selection~\cite{radford2021learning} yields only marginal gains, improving accuracy from 29.6\% to 30.6\%, while manually selecting informative frames raises accuracy to 36.1\%. These results suggest that VLMs need a more effective way to organize and retrieve spatial evidence from redundant visual inputs.

Frame selection only filters existing observations, it cannot construct a geometrically consistent scene representation or expose spatial information from viewpoints that are not directly observed. Recent methods such as MindJourney~\cite{yang2026mindjourney} address this limitation by generating additional views, but generated views may lack the spatial consistency needed for reliable reasoning. In contrast, reconstruction-based Vision Foundation Models (VFMs) offer a natural way to aggregate multi-view images or videos into explicit spatial memory, such as point clouds. This memory preserves geometric structure and supports viewpoint transformation, rendering, and spatial querying~\cite{wang2025vggt,wang2026pi}.

\begin{figure}[t]
  \centering
   \includegraphics[width=0.9\linewidth]{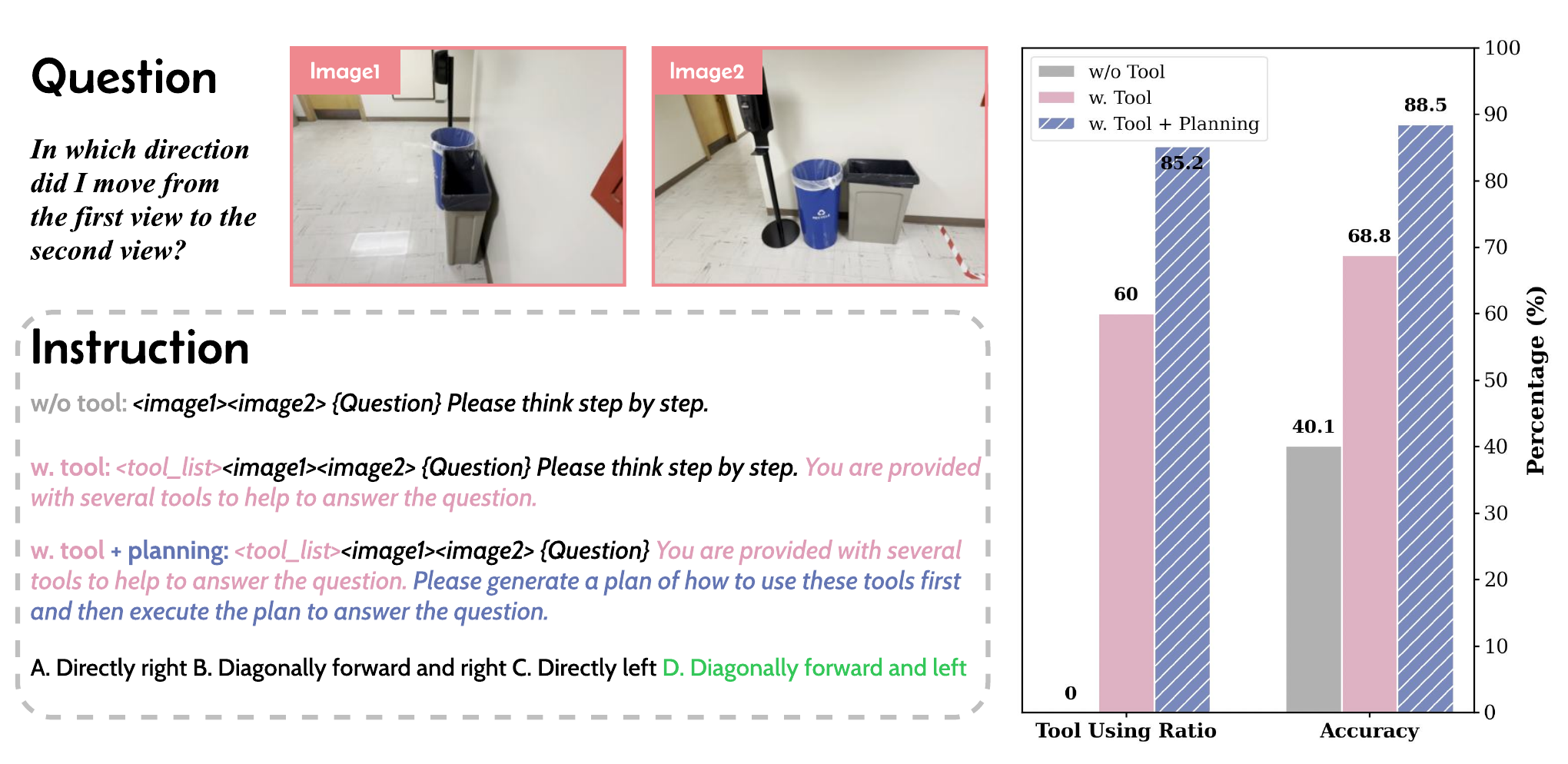}
   \vspace{-2mm}
   \caption{Camera-transition results on MindCube. Explicit tool-use planning yields more reliable performance than direct tool access.}
   \label{fig:motivation}
   \vspace{-5mm}
\end{figure}

However, effective spatial reasoning also depends on how VLMs interact with this memory. Simply exposing reconstruction models as free-form tools is brittle: VLMs may invoke tools incorrectly, skip required spatial transformations, or misuse intermediate results. For example, as shown in Figure~\ref{fig:motivation}, on camera-transition tasks from MindCube~\cite{yin2025spatial}, GPT-5-mini improves by more than 20 percentage points with direct tool access, but remains unstable because it uses the provided tools in only 60\% of cases. In contrast, explicit tool-use planning raises the tool-use rate to 83.2\% and improves accuracy by an additional 20 percentage points. These observations motivate structured, verifiable interaction with spatial memory.

Based on these insights, we propose \textbf{Reasmory} (\textbf{Re}construction \textbf{as} Me\textbf{mory}), a framework that formulates spatial reasoning as structured program execution over reconstructed spatial memory. As shown in Figure~\ref{fig:teaser}, Reasmory first constructs explicit 3D memory from multi-view images or videos using VFMs and augments it with semantically grounded 3D object instances. It then introduces a lightweight Domain-Specific Language (DSL) that constrains how VLMs query objects and cameras, transform viewpoints, and render observations during reasoning. Generated programs are parsed and validated for syntactic correctness, function usage, and execution dependencies before deterministic execution, after which a reasoner produces the final answer.
This design turns spatial reasoning into a controlled sequence of memory operations rather than unconstrained tool calls. We evaluate Reasmory on multi-view image and video spatial reasoning benchmarks, achieving consistent gains of 6--18\% over strong baselines, including GPT-5-mini and Gemini-3-flash.

Our contributions are threefold. First, we propose \textbf{Reasmory}, a framework that constructs explicit 3D spatial memory, augments it with semantically grounded 3D object instances, and formulates spatial reasoning as program execution over reconstructed spatial memory. Second, we introduce a lightweight DSL with validation mechanisms that constrain querying, viewpoint transformation, and rendering over spatial memory, improving robustness over free-form tool use. Third, we evaluate Reasmory on multi-view image and video spatial reasoning benchmarks and show improvements over strong VLM and test-time scaling baselines.

\section{Related Work}
\noindent \textbf{Spatial Memory.}
Spatial cognition studies suggest that structured memory enables viewpoint transformation and navigation~\cite{tolman1948cognitive,siegel1975development,o1978hippocampus,o1987single}. Inspired by this, modern models incorporate memory to improve spatial consistency. In streaming 3D reconstruction, memory is maintained either implicitly through hidden states~\cite{wang20253d,wang2025continuous,chen2026tttr,zhang2026loger,zhuo2026streaming} or explicitly through feature-augmented 3D points~\cite{wu2026pointr}. Similar trends appear in video generation, where models use implicit memory~\cite{po2025long,savov2026statespacediffuser,zhang2026testtime} or explicit mechanisms such as retrieval and 3D point representations~\cite{yu2025context,li2025vmem,xiao2026worldmem,zhou2025learning,liu2025dynamem}. In embodied reasoning, memory is often represented as token sequences or graphs~\cite{he2025mem4nav,liu2026msnav,zhang2025constructing,zemskova20253dgraphllm}. In contrast, Reasmory constructs explicit 3D spatial memory at inference time, augments it with semantically grounded 3D object instances, and exposes operations that can be queried, transformed, rendered, and validated.

\noindent \textbf{3D Reconstruction and Spatial VLMs.}
Advances in NeRF~\cite{mildenhall2021nerf,chen2022tensorf,muller2022instant,martin2021nerf}, Gaussian Splatting~\cite{kerbl20233d,you2025instainpaint}, and feed-forward reconstruction~\cite{zhang2026loger,wang2024dust3r,wang2025vggt,wang2026pi,lin2026depth} have made efficient 3D geometry estimation practical. Recent spatial VLMs inject geometric structure through spatial data generation and supervision~\cite{chen2024spatialvlm,ouyang2025spacer,feng2026videor}, geometry or reconstruction encoders~\cite{wu2026spatialmllm,fan2025vlm,hu2025g2vlm,zhao2025spacemind}, or distilled 3D-aware features~\cite{huang20253daware,chen2025think}. These methods improve spatial understanding through model training or feature integration. Other approaches operate at test time by generating additional views~\cite{yang2026mindjourney}, invoking 3D tools~\cite{zhang2026think3d,luo2026pyspatial}, or encoding geometric references for an MLLM~\cite{yuan2026gr3d}. Reasmory is closest to this line of test-time methods, but differs by turning reconstruction-based memory access into validated DSL program execution rather than unconstrained tool invocation.

\noindent \textbf{Programmatic Tool Use and DSLs.}
Tool-augmented reasoning frameworks such as ReAct~\cite{yao2023react} and Toolformer~\cite{schick2023toolformer} show
that tools can improve reasoning, while program-aided approaches delegate parts of reasoning to executable
code~\cite{gao2022pal}. In visual reasoning, systems such as VISPROG~\cite{gupta2022visprog} and
ViperGPT~\cite{suris2023vipergpt} generate programs that compose vision modules at inference time. Recent spatial
agents such as MSSR~\cite{guo2026mssr} query 3D scenes with expert tools and prune redundant spatial information
before answering. These methods show the value of external computation, but executable interaction with vision and 3D
tools can still be brittle in spatial tasks, where errors in viewpoint state, tool order, or intermediate outputs can
corrupt reasoning. Domain-specific languages (DSLs) provide structured and verifiable intermediate
representations~\cite{mernik2005and,fowler2010domain,ellis2021coder}. Building on this paradigm, Reasmory introduces a
lightweight spatial DSL that validates generated programs before execution over explicit 3D memory, directly
controlling tool usage, dependencies, and viewpoint-state transitions.

\section{Method}
\begin{figure}[t]
  \centering
   \includegraphics[width=\linewidth]{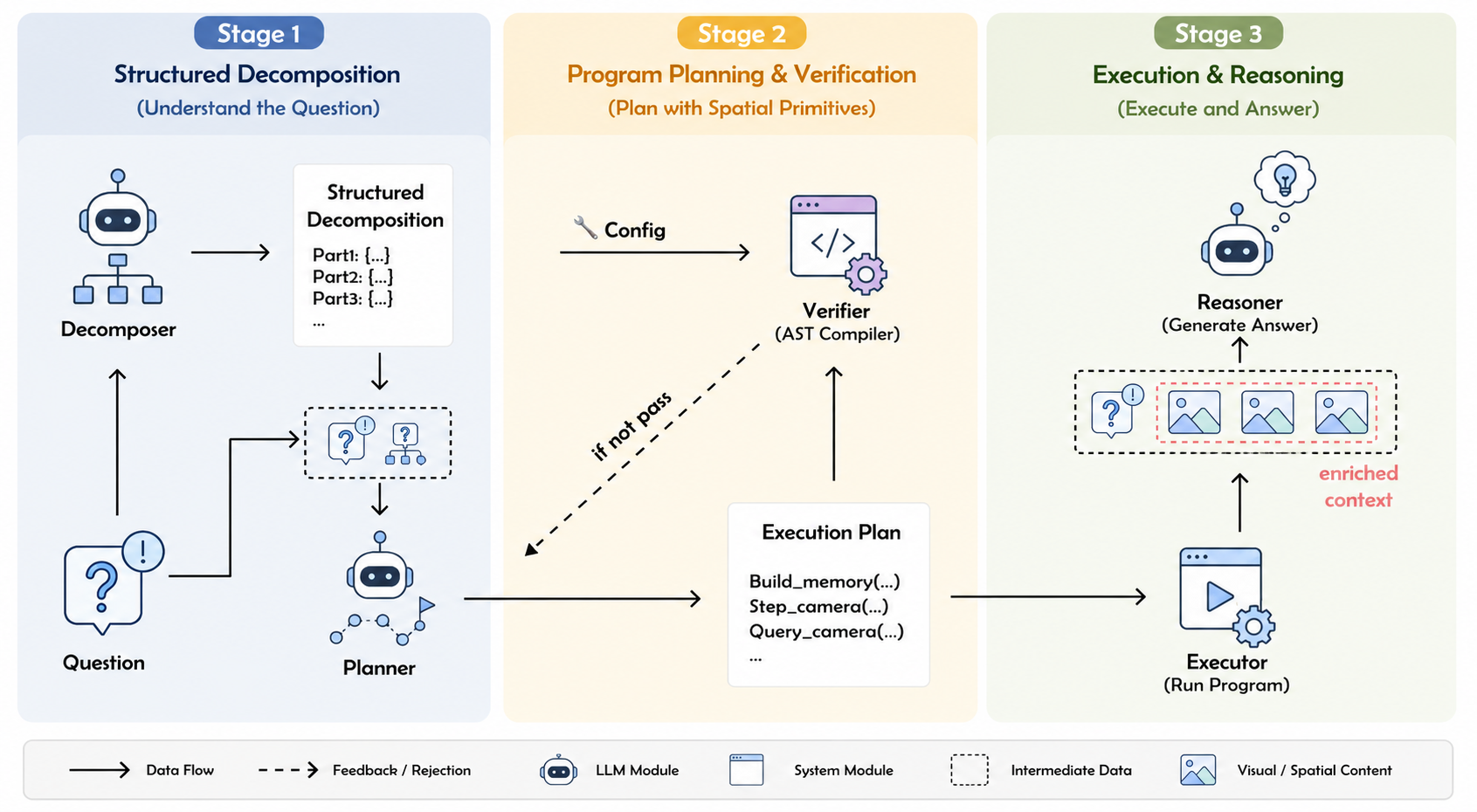}
   \vspace{-5mm}
  \caption{Overview of \textbf{Reasmory}. The system constructs reconstruction-based spatial memory, augments it with grounded 3D object instances, generates and validates a DSL program, and executes the program to support spatial reasoning.}
   \label{fig:pipeline}
   \vspace{-5mm}
\end{figure}
Reasmory addresses two key challenges: organizing spatial information into an explicit memory, and controlling how VLMs access and manipulate that memory during reasoning. 
As shown in Figure~\ref{fig:pipeline}, Reasmory first constructs reconstruction-based 3D memory from multi-view images or video observations and augments it with grounded 3D object instances. It then answers spatial questions by generating, validating, and executing DSL programs over this memory.
The remainder of this section is organized as follows: Sec.~\ref{sec:prelim} introduces feed-forward 3D reconstruction and DSLs; Sec.~\ref{sec:recon} presents spatial memory construction and semantic augmentation; and Sec.~\ref{sec:interaction} explains how Reasmory decomposes each question, plans a DSL program with spatial primitives, then validates the program and executes it over spatial 
memory.

\begin{algorithm}[t]
\scriptsize
\caption{{Semantic augmentation with multi-view 3D agreement}}
\label{alg:semantic_augmentation}
\begin{algorithmic}[1]
\Require Views $\{V_i\}_{i=1}^{n}$, reconstructed 3D maps $\{\mathbf{X}^{(i)}\}_{i=1}^{n}$, query categories $\mathcal{C}$
\Statex \textbf{Output:} 3D object instances $\mathcal{O}$ with category labels, centers, and associated 3D points
\State $\mathcal{O} \gets \emptyset$
\ForAll{category $c \in \mathcal{C}$}
    \State Extract 2D segmentation proposals for $c$ in all views
    \State Remove duplicate proposals within each view
    \ForAll{remaining proposal mask $M$ in view $i$}
        \State Lift $M$ into 3D using $\mathbf{X}^{(i)}$
        \State Store the occupied 3D grid cells for this proposal
        \State Compare these cells with same-category proposals in other views
        \State Score $M$ by its cross-view 3D agreement
    \EndFor
    \State Keep proposals with sufficient agreement
    \State Build a graph connecting mutually consistent proposals
    \State Greedily merge connected proposals, with at most one mask per view
    \State Add each merged cluster as a 3D object instance in $\mathcal{O}$
\EndFor
\end{algorithmic}
\end{algorithm}
\begin{algorithm}[t]
\caption{{Spatial reasoning decomposition. $T_V(\mathcal{E})$ are entities expressed in viewpoint $V$}}
\scriptsize
\label{alg:decomposition}
\begin{algorithmic}[1]
\Require Question $q$
\State $V_{\mathrm{ini}} \gets \textsc{SelectReference}(q)$
\State $V \gets V_{\mathrm{ini}}$
\While{$\neg\,\textsc{ReadyForComparison}(V, q)$}
    \State $V \gets \textsc{RefineViewpoint}(V, q)$
\EndWhile
\State $\mathcal{E} \gets \textsc{IdentifyEntities}(q)$
\State $a \gets f\!\left(T_V(\mathcal{E})\right)$
\Return $a$
\end{algorithmic}
\end{algorithm}

\begin{table*}[t]
    \centering
    \scriptsize
    \setlength{\tabcolsep}{4pt}
    \caption{
    Examples of spatial reasoning decomposition. Each case selects a reference viewpoint when needed, applies optional viewpoint refinement, and then compares entities under the transformed viewpoint.
    }
    \begin{tabular}{p{0.36\textwidth} p{0.58\textwidth}}
    \toprule
    \textbf{Problem} & \textbf{Decomposition} \\
    \midrule
    
    How many tables are in the room? 
    &
    \textbf{Reference:} none.
    \newline
    \textbf{Refinement:} none.
    \newline
    \textbf{Comparison:} count(table).
    \newline
    \textbf{Primitives:} Build $\rightarrow$ Query(table) $\rightarrow$ BEV(table).
    
    \\
    \midrule
    
    In which direction did I move from the first view to the second view?
    &
    \textbf{Reference:} $V_{\text{ini}} =$ camera$_1$.
    \newline
    \textbf{Refinement:} none.
    \newline
    \textbf{Comparison:} direction(ego, camera$_2$).
    \newline
    \textbf{Primitives:} Build $\rightarrow$ Query(cam$_1$) $\rightarrow$ SetView(cam$_1$) $\rightarrow$ Query(cam$_2$) $\rightarrow$ BEV(cam$_2$).
    
    \\
    \midrule
    What is behind me if I stand at the same spot and facing direction as image 3?
    &
    \textbf{Reference:} $V_{\text{ini}} =$ camera$_3$.
    \newline
    \textbf{Refinement:} rotate(back).
    \newline
    \textbf{Comparison:} visibility(ego, object).
    \newline
    \textbf{Primitives:} Build $\rightarrow$ Query(cam$_3$) $\rightarrow$ SetView(cam$_3$) $\rightarrow$ Turn(back) $\rightarrow$ RGB.
    
    \\
    \midrule

    If I am standing at the same spot and facing the same direction as shown in image 1, then I turn right and move forward, will I get closer to the pink plush toy and headboard?
    &
    \textbf{Reference:} $V_{\text{ini}} =$ camera$_1$.
    \newline
    \textbf{Refinement:} rotate(right), move(forward).
    \newline
    \textbf{Comparison:} direction(ego, plush toy).
    \newline
    \textbf{Primitives:} Build $\rightarrow$ Query(cam$_1$) $\rightarrow$ SetView(cam$_1$) $\rightarrow$ Turn(right) $\rightarrow$ Step(forward) $\rightarrow$ RGB.
    
    \\
    
    \bottomrule
    \end{tabular}
    \label{tab:decomposition}
    \vspace{-3mm}
\end{table*}
\begin{table*}[t]
\centering
\begin{minipage}[t]{\textwidth}
\centering
\scriptsize
\caption{
Spatial reasoning primitives and stage-specific constraints that together define valid interactions between the planner and explicit spatial memory.
}
\begin{tabular}{p{3cm} p{8.7cm}}
\toprule
\textbf{Primitive Type} & \textbf{Operations} \\
\midrule
\textbf{Memory Construction} & \texttt{build\_static\_memory}, 
\texttt{build\_dynamic\_memory}
\\
\textbf{Memory Query} & \texttt{query\_camera\_pose}, \texttt{query\_3d\_object\_location}
\\
\textbf{Memory Transformation} & \texttt{set\_viewpoint}, \texttt{step\_camera}, \texttt{turn\_camera}
\\
\textbf{Rendering} & \texttt{render\_egocentric}, \texttt{render\_semantic\_bev}, \texttt{render\_rgb\_bev}
\\
\bottomrule
\end{tabular}
\vspace{1mm}

{\footnotesize (a) Spatial reasoning primitives.}
\vspace{2mm}
\end{minipage}
\hfill
\begin{minipage}[t]{\textwidth}
\centering
\scriptsize
\begin{tabular}{p{3.7cm} p{6.0cm}}
\toprule
\textbf{Reasoning Stage} & \textbf{Allowed Primitive Types} \\
\midrule
Reference Viewpoint Selection & Memory Construction, Memory Query
\\
Viewpoint Refinement & Memory Query, Memory Transformation, Rendering
\\
Entity Comparison & Memory Query, Rendering
\\
\bottomrule
\end{tabular}
\vspace{1mm}

{\footnotesize (b) Allowed primitives for each reasoning stage.}
\end{minipage}
\label{tab:primitives}
\end{table*}

\subsection{Preliminaries}
\label{sec:prelim}
We briefly introduce the two foundations of Reasmory: feed-forward 3D reconstruction and domain-specific languages.

\noindent \textbf{Feed-forward 3D Reconstruction.}
Given multi-view images or frames sampled from a video, denoted by $\{V_1, V_2, \dots, V_n\}$, a feed-forward 3D reconstructor $\mathcal{R}$ estimates dense depth maps $\{D_1, D_2, \dots, D_n\}$, camera poses $\{T_1, T_2, \dots, T_n\}$, and camera intrinsics $K$ in a single forward pass.
Using the predicted depth and camera parameters, each pixel can be back-projected into 3D. Specifically, for a pixel $p=(u,v)$ in image $V_i$ with depth value $D_i(u,v)$, the corresponding 3D point is computed as:
\begin{equation}
    \mathbf{x}
=
T_i^{-1}
\left(
D_i(u,v)\cdot K^{-1}
\begin{bmatrix}
u \\ v \\ 1
\end{bmatrix}
\right).
\end{equation}
Each 3D point is associated with its RGB value from the input image, yielding a colored point cloud representation of the scene.

\noindent \textbf{Domain-Specific Language.}
A Domain-Specific Language (DSL)~\cite{mernik2005and} defines a restricted programming interface for a specific task domain. A DSL typically consists of: (1) a syntax that defines valid operations and program structures; (2) an Abstract Syntax Tree (AST), which represents program structure and dependencies; and (3) a compiler. In our setting, the DSL specifies the valid spatial-memory operations, the program structure used to compose them, and the compiler checks applied to the resulting AST before execution.
Compared with free-form tool use, this provides a more controllable and verifiable interface for reasoning over spatial memory.



\subsection{Spatial Memory Construction}
\label{sec:recon}

\textbf{Memory Representation.} 
Reasmory represents each scene as an explicit 3D spatial memory in the form of a colored point cloud. Given multi-view images or video frames, we reconstruct a colored point cloud using a feed-forward reconstructor $\mathcal R$ (e.g., Pi3~\cite{wang2026pi} or Depth Anything v3~\cite{lin2026depth}). The memory stores the point cloud $\mathbf X$, camera trajectories $\mathbf T$, and depth maps $\mathbf D$ for all input frames. This representation fuses unordered, partially overlapping observations into a geometrically consistent scene structure while preserving information needed for viewpoint transformation, rendering, and spatial querying.
To simplify downstream spatial reasoning, we align the memory to canonical axes. Specifically, we align the global up direction with the $y$-axis and use minimum-area bounding-box alignment to align dominant wall directions with the $x$- and $z$-axes, yielding a more interpretable coordinate system and making the problems related to size estimation easier.

\noindent \textbf{Semantic Augmentation.}
Many spatial reasoning questions refer to objects by category or name, such as chairs, tables, or other named objects. To support such object-level queries, we augment the reconstructed 3D memory with semantically grounded object instances. Each instance stores a category label, an estimated 3D center, and the set of associated 3D points, enabling operations such as \texttt{query\_3d\_object\_location}.
Algorithm~\ref{alg:semantic_augmentation} summarizes the construction process. For each queried category, we extract category-specific segmentation proposals in all views using SAM3~\cite{carion2025sam}, lift each proposal into 3D using the reconstructed geometry, and represent it by occupied 3D grid cells. Proposals corresponding to the same physical object should occupy consistent 3D regions across views. We therefore score each lifted proposal by how consistently its occupied grid cells match same-category proposals in other views. Proposals with sufficient agreement are connected in a category-specific graph and greedily merged under a one-instance-per-view constraint. Scoring thresholds and implementation details are provided in the supplementary material.

\subsection{Structured Interaction with Spatial Memory}
\label{sec:interaction}

As illustrated in Figure~\ref{fig:pipeline}, the interaction pipeline consists of three stages. First, a \emph{Question Decomposer} analyzes the spatial reasoning problem and converts it into a structured representation. Second, a \emph{Planner} generates a DSL program composed of spatial reasoning primitives that describes how to interact with the spatial memory. To ensure correctness, the generated program is checked by an AST-based compiler and iteratively refined using compiler feedback when necessary. Third, a validated program is executed over the spatial memory to retrieve additional observations and spatial evidence, which are then incorporated into the context for the final \emph{Reasoner}. Together, they transform spatial reasoning from unconstrained tool usage into a verifiable interaction process with explicit spatial memory.

\noindent \textbf{Question Decomposition.}
The decomposer maps each spatial reasoning question into three stages. First, \emph{reference viewpoint selection} chooses an initial viewpoint $V_{\mathrm{ini}}$ when the question is anchored to a specific image, camera, or observer pose. Second, \emph{viewpoint refinement} updates this viewpoint through rotations or translations specified by the question. Third, \emph{entity transformation and comparison} identifies the relevant entities, expresses them under the refined viewpoint, and applies a task-specific comparison function $f$, such as direction, distance, visibility, or counting. 
This decomposition provides a structured scaffold for planning.
Algorithm~\ref{alg:decomposition} summarizes this abstraction, and Table~\ref{tab:decomposition} provides representative examples.

\noindent \textbf{Program Planning with Spatial Primitives.}
Given this decomposition, the planner generates a DSL program using atomic spatial reasoning primitives. These primitives fall into four categories: \emph{memory construction}, which builds or loads the spatial memory; \emph{memory query}, which retrieves camera or object information; \emph{memory transformation}, which updates the current viewpoint; and \emph{rendering}, which produces egocentric or bird's-eye-view observations for visual inspection.
We design the DSL as a restricted subset of Python, where the planner can compose only these allowed primitives and must follow the program structure checked by the compiler described below.
Different reasoning stages are associated with different allowable primitive categories. For example, viewpoint refinement permits memory transformation and rendering operations, while reference viewpoint selection only allows memory construction and querying. This constrained interaction design reduces invalid reasoning trajectories and enforces consistent state transitions during execution. Detailed primitive categories and their corresponding stage constraints are summarized in Table~\ref{tab:primitives}.

\noindent \textbf{Program Validation and Execution.}
\label{sec:dsl}
\begin{table}[t]
\centering
\scriptsize
\setlength{\tabcolsep}{4pt}
\caption{Representative AST validation rules in the main text, organized by taxonomy category. The full taxonomy is deferred to the appendix.}
\begin{tabular}{p{0.28\linewidth} p{0.28\linewidth} p{0.34\linewidth}}
\toprule
\textbf{Category} & \textbf{Representative rule} & \textbf{Description} \\
\midrule
Program syntax
& Restricted Python subset
& Decorators, default arguments, variadic arguments, keyword-only arguments, and return annotations are disallowed. \\
\midrule
Tool usage
& Whitelisted primitives
& Only predefined spatial memory primitives in the DSL may be called. \\
\midrule
Dependency consistency
& Define-before-use
& Variables must be defined before they are referenced. \\
\midrule
Viewpoint-state consistency
& Stale-query invalidation
& Query results derived under a previous viewpoint become invalid after viewpoint-changing operations and must be recomputed. \\
\midrule
Execution discipline
& No dangling motion
& Each camera-motion operation must be followed by at least one subsequent render call. \\
\midrule
Plan consistency
& Motion-direction alignment
& Camera motions must exactly match the directions specified by the decomposition. \\
\bottomrule
\end{tabular}
\label{tab:ast-taxonomy-summary}
\end{table}
The program generated by the planner is passed to an \emph{AST Compiler}, which verifies correctness before execution. As shown in Table~\ref{tab:ast-taxonomy-summary}, the compiler checks: (1) syntactic validity; (2) valid function usage; (3) dependency consistency; (4) viewpoint state consistency; (5) execution discipline; and (6) plan consistency. Programs that fail validation are rejected and regenerated using compiler feedback. Once validated, the \emph{Execution Engine} runs the program deterministically over the spatial memory, producing query results, transformed viewpoints, and rendered observations. Finally, the \emph{Reasoner} uses these execution outputs, together with the original question, to produce the final answer.
This converts spatial reasoning from unconstrained tool use into validated program execution, where invalid operations, missing dependencies, and inconsistent state transitions can be 
detected before execution.
Compiler feedback also enables iterative self-correction, allowing the planner to repair most invalid programs with only a few regeneration attempts.

\section{Experiments}

We evaluate Reasmory on three input modalities that stress complementary forms of evidence: multi-view images from MindCube~\cite{yin2025spatial} with a small set of observed views, static-scene videos from VSI-Bench~\cite{yang2025thinking} with cues distributed across redundant frames, and dynamic-scene videos from VLM4D~\cite{zhou2025vlm4d} with both camera and object motion. We compare against direct inference with GPT-5-mini and Gemini-3-flash, spatial QA fine-tuned models~\cite{wu2026spatialmllm,li2026spatialladder,VQASynth}, and test-time scaling methods including MindJourney~\cite{yang2026mindjourney} and Think3D~\cite{zhang2026think3d}. Sec.~\ref{sec:exp_setup} details the setup, Sec.~\ref{sec:exp_main} presents benchmark comparisons, Sec.~\ref{sec:exp_ablation} ablates DSL verification, Sec.~\ref{sec:exp_reliability} analyzes semantic object grounding and planner validity, and Sec.~\ref{sec:exp_example} provides an end-to-end Reasmory reasoning example.

\subsection{Experimental Setup}
\label{sec:exp_setup}
\noindent \textbf{Vision Modules for Spatial Memory.} Reasmory uses external vision modules to construct spatial memory and support semantic object queries. Specifically, we use Pi3~\cite{wang2026pi} to reconstruct point clouds, camera poses, and depth maps from input image sequences. In addition, the metric-depth variant Pi3X is used when the agent needs to estimate object sizes at real-world scale. For experiments on dynamic-scene videos, we use Flow3r~\cite{cong2026flow3r} to estimate geometry and camera poses across video frames. All images used for reconstruction are downsampled to a resolution with the short edge set to 378 pixels. To better utilize video inputs, we sample denser frame sequences for Pi3 reconstruction (\eg, 64 frames), while using sparser inputs for VLM reasoning (\eg, 16 frames) to reduce visual redundancy and accelerate reasoning. For the \texttt{query\_3d\_object\_location} operation, we use SAM3~\cite{carion2025sam} to perform semantic segmentation. For video inputs, SAM3 processes all 64 frames to support cross-view merging. To ensure higher precision, we set the confidence threshold to 0.65.

\noindent \textbf{Benchmarks.} We evaluate our pipeline on three spatial reasoning benchmarks covering multi-view images, static-scene videos, and dynamic-scene videos. For multi-view reasoning, we use MindCube~\cite{yin2025spatial} and uniformly sample 50 problems from each of the \textit{Among}, \textit{Rotation}, and \textit{Around} categories, resulting in a 150-sample evaluation set, MindCube-Tiny. We observe that the original MindCube benchmark contains strong textual cues for spatial reasoning, such as:
\emph{``Based on these four images (image 1, 2, 3, and 4) showing the blue table from different viewpoints (front, left, back, and right), ...''}
Such descriptions leak viewpoint information by explicitly annotating the camera pose of each image. This substantially reduces the need for cross-view correspondence reasoning and introduces a shortcut that may overestimate a model's true spatial reasoning capability.
To study this effect, we remove textual hints and shuffle both image order and answer options when evaluating GPT-5-mini and Gemini-3-flash. As shown in Table~\ref{tab:debias}, Gemini-3-flash drops by more than 5\% after removing textual hints and by an additional 6\% after shuffling image order; GPT-5-mini follows the same trend. We therefore use this debiased MindCube-Tiny split for subsequent experiments. For static-scene videos, we sample 50 problems from each VSI-Bench~\cite{yang2025thinking} category, yielding 400 questions. For dynamic scenes, we sample 50 examples from both the ego-centric and exo-centric splits of VLM4D~\cite{zhou2025vlm4d}.

\noindent \textbf{Models.} Reasmory is a test-time framework that can be applied to different backbone VLMs. We evaluate it with two strong frontier models, GPT-5-mini and Gemini-3-flash, to test whether structured spatial memory can improve already capable multimodal reasoners. We access both models through APIs using the same default temperature and reasoning settings across all benchmarks. For comparison, we include two groups of baselines. The first group consists of spatial QA fine-tuned models, including SpatialMLLM~\cite{wu2026spatialmllm}, SpatialLadder~\cite{li2026spatialladder}, and SpaceOM~\cite{VQASynth}. The second group consists of test-time scaling methods: MindJourney~\cite{yang2026mindjourney}, which uses video generation models to simulate camera motion, and Think3D~\cite{zhang2026think3d}, which also leverages 3D reconstruction for reasoning but without explicit primitive design or constrained interaction. For different agents in our pipeline, we instantiate multiple copies of the same backbone model, assigning them specialized roles using different prompts. 

\begin{table}[t]
\centering
\scriptsize
\setlength{\tabcolsep}{4pt}
\caption{Debiasing analysis on MindCube-Tiny. Removing textual viewpoint hints and shuffling image order both reduce performance, suggesting that the original benchmark leaks camera-pose information and partially bypasses cross-view correspondence reasoning.}
\begin{tabular}{lccc}
\toprule
 & \textbf{Mindcube-Tiny} & \textbf{+Remove Hint} & \textbf{+Shuffle Image Order} \\
\midrule
\textbf{GPT-5-mini} & \textbf{59.4} & 58.6 (-0.8) & 58.0 (-1.4)\\
\textbf{Gemini-3-flash} & \textbf{81.9} & 74.6 (-7.3) & 68.7 (-13.2) \\
\bottomrule
\end{tabular}
\label{tab:debias}
\vspace{-1em}
\end{table}
\begin{table}[t]
\centering
\scriptsize
\caption{\textbf{Evaluation results on de-biased MindCube-Tiny.} We compare methods under each base model separately. 
Bold numbers indicate the best result within the same base model family, and underlined numbers indicate the best result overall.
}
\label{tab:eval-mindcube}
\vspace{1mm}
\begin{tabular}{lcccc}
\toprule
\textbf{Method} & \textbf{Among $\uparrow$}& \textbf{Rotation $\uparrow$}& \textbf{Around $\uparrow$} & \textbf{Overall $\uparrow$} \\
\midrule
GPT-5-mini & 32.0 & 78.0 & 64.0 & 58.0\\
Gemini-3-flash & 56.0 & 70.0 & 74.0 & 68.7\\
\midrule
\multicolumn{2}{l}{\textit{\textbf{Spatial Models}}}
\\
\midrule
SpaceOM-3B~\cite{VQASynth} & 38.0 & 32.0 & 62.0 & 44.0\\
Spatial-MLLM-4B~\cite{wu2026spatialmllm} & 47.5 & 32.5 & 35.0 & 38.3 \\
SpatialLadder-3B~\cite{li2026spatialladder} & 50.0 & 40.0 & 56.0 & 48.7\\
\midrule
\multicolumn{2}{l}{\textit{\textbf{Test-time Scaling Methods}}}\\
\midrule
Mindjourney (GPT-5-mini)~\cite{yang2026mindjourney} & 36.0 & 52.0 & 74.0& 54.0 \textcolor{red}{\textbf{(-4.0)}} \\
Mindjourney (Gemini-3-flash)~\cite{yang2026mindjourney} & 50.0 & 74.0& 74.0& 66.0 \textcolor{red}{(-2.7)}\\
Think3D (GPT-5-mini)~\cite{zhang2026think3d} & 54.0 & 56.0 & 62.0 & 56.7 \textcolor{red}{(-1.3)}\\
Think3D (Gemini-3-flash)~\cite{zhang2026think3d} & 54.0 & 66.0 & 82.0 & 67.3 \textcolor{red}{(-1.4)}\\
\rowcolor{cyan!15}
\textbf{Reasmory (GPT-5-mini)} & \textbf{78.0} & 76.0 & \textbf{74.0}& \textbf{76.0 (+18.00)}\\
\rowcolor{cyan!15}
\textbf{Reasmory (Gemini-3-flash)} & \underline{\textbf{82.0}} & \underline{\textbf{92.0}} & \underline{\textbf{84.0}} & \underline{\textbf{86.0 (+17.3)}}\\
\bottomrule
\end{tabular}
\vspace{-3mm}
\end{table}

\subsection{Benchmark Results}
\label{sec:exp_main}

\noindent \textbf{Multi-view Images.} Table~\ref{tab:eval-mindcube} presents results on the debiased MindCube-Tiny benchmark, where each sample contains 2--4 input images. Existing spatially fine-tuned models~\cite{wu2026spatialmllm,li2026spatialladder,VQASynth} are not competitive with frontier closed-source VLMs. For example, Spatial-MLLM achieves only 38.3\% accuracy. We hypothesize that these relatively small-scale models (3B/4B) struggle to generalize to the diverse spatial configurations in MindCube, which may differ from their training distributions. Among test-time scaling approaches, existing methods improve only specific subsets of problems. For example, MindJourney improves GPT-5-mini by 10 points on \textit{Around} problems, while Think3D improves GPT-5-mini by 22 points on \textit{Among} problems. However, these gains do not generalize across categories, leading to limited or even negative overall improvements. In contrast, Reasmory substantially improves performance for both frontier models. GPT-5-mini improves from 58.0\% to 76.0\% (+18.0), and Gemini-3-flash improves from 68.7\% to 86.0\% (+17.3). Reasmory also achieves balanced performance across \textit{Among}, \textit{Rotation}, and \textit{Around} tasks, suggesting that explicit spatial memory and constrained program execution provide a more general mechanism for multi-view spatial reasoning than existing test-time scaling approaches.

\noindent \textbf{Video-based Static Scenes.} Table~\ref{tab:eval-vsi} presents results on VSI-Bench-Tiny, a video-based spatial reasoning benchmark for static scenes. Compared with de-biased MindCube-Tiny, all methods achieve lower overall accuracy, and direct frontier VLM inference remains around 50\%. This suggests that long video inputs introduce substantial visual redundancy and make spatial evidence harder to retrieve. The gap between spatially fine-tuned models and frontier VLMs is also smaller than on MindCube-Tiny. This may be because VSI-Bench is closer to the training distributions of spatial models, such as Spatial-MLLM-120K~\cite{wu2026spatialmllm} and SpatialLadder-26K~\cite{li2026spatialladder}. In contrast, SpaceOM~\cite{VQASynth}, which is primarily trained in single-image dataset SpaceThinker, fails to generalize to video-based reasoning tasks. 

Among test-time scaling approaches, MindJourney~\cite{yang2026mindjourney} performs poorly in video settings. Since the original method is designed for two-image inputs, we adapt it by using one frame as the anchor image and stitching the remaining 16 frames into a $4 \times 4$ grid as the helper input. The resulting performance degradation suggests that generated-view reasoning discards substantial information from long video inputs, limiting its effectiveness for long-context spatial reasoning.
For Think3D~\cite{zhang2026think3d}, although the method provides access to 3D reconstruction tools, performance on several sub-tasks becomes worse than direct free-form reasoning. We attribute this to unstable interaction with reconstruction tools, indicating that unconstrained tool usage may confuse the model during reasoning.
In contrast, Reasmory achieves the best overall performance for both frontier backbones. It improves GPT-5-mini from 49.6\% to 55.8\% (+6.2) and Gemini-3-flash from 51.6\% to 65.0\% (+13.4). The gains are especially strong on geometry-intensive tasks such as \textit{Absolute Distance}, \textit{Relative Distance}, and \textit{Room Size}, demonstrating that explicit spatial memory and constrained program execution are particularly effective for long-context spatial reasoning in video settings.
\begin{table}[t]
\centering
\scriptsize
\setlength{\tabcolsep}{2pt}
\renewcommand{\arraystretch}{1.1}
\caption{
\textbf{Evaluation results on video-input VSI-Bench-Tiny.} We compare methods under each base model separately. 
Bold numbers indicate the best result within the same base model family, and underlined numbers indicate the best result overall.
}
\label{tab:eval-vsi}
\vspace{1mm}

\begin{tabular}{lccccccccc}
\toprule

\textbf{Method}
& \rotatebox{50}{\textbf{Obj. Count}}
& \rotatebox{50}{\textbf{Abs. Dist.}}
& \rotatebox{50}{\textbf{Obj. Size}}
& \rotatebox{50}{\textbf{Room Size}}
& \rotatebox{50}{\textbf{Rel. Dist.}}
& \rotatebox{50}{\textbf{Rel. Dir.}}
& \rotatebox{50}{\textbf{Route Plan}}
& \rotatebox{50}{\textbf{Appr. Order}}
& \rotatebox{50}{\textbf{Overall}}
\\
\midrule

GPT-5-mini 
& 49.0 & 31.6 & 73.0 & 41.2 & 42.0 & 48.0 &\textbf{ 46.0} & 66.0 & 49.6 \\

Gemini-3-flash (16 frames) 
& 46.4 & 15.2 & 68.6 & 48.4 & 56.0 & 58.0 & 48.0 & 72.0 & 51.6 \\

Gemini-3-flash (video) 
& 45.9 & 21.8 & \underline{\textbf{75.0}} & 39.2 & 58.0 & 62.0 & 40.0 & \underline{\textbf{88.0}} & 53.7 \\

\midrule

\multicolumn{10}{l}{\textit{\textbf{Spatial Models}}}
\\
\midrule
SpaceOM-3B~\cite{VQASynth}
& 27.3 & 10.2 & 17.1 & 21.7 & 10.2 & 25.0 & 22.4 & 31.9 & 20.7 \\
Spatial-MLLM-4B~\cite{wu2026spatialmllm} 
& \textbf{67.2} & 31.0 & 56.6 & 43.6 & 32.0 & 52.0 & 36.0 & 44.0 & 45.2 \\
SpatialLadder-3B~\cite{li2026spatialladder}
& 60.0 & 31.0 & 54.2 & 47.6 & 30.0 & 44.0 & 30.0 & 40.0 & 42.1 \\

\midrule

\multicolumn{10}{l}{\textit{\textbf{Other Test-time Scaling Methods}}}
\\
\midrule

Mindjourney (GPT-5-mini)~\cite{yang2026mindjourney}
& 47.0 & 25.0 & 64.5 & 23.0 & 20.0 & 50.0 & 30.0 & 50.0 & 38.7 \textcolor{red}{\textbf{(-10.9)}} \\

Mindjourney (Gemini-3-flash)~\cite{yang2026mindjourney}
& 25.5 & 36.0 & 59.5 & 56.5 & 35.0 & 50.0 & \underline{\textbf{65.0}} & 80.0 & 50.9 (-0.7) \\

Think3D (GPT-5-mini)~\cite{zhang2026think3d}
& 42.4 & 25.2 & 63.8 & 41.2 & 34.0 & 48.0 & 31.0 & 71.7 & 44.7 \textcolor{red}{(-4.9)} \\

Think3D (Gemini-3-flash)~\cite{zhang2026think3d}
& 45.8 & 10.2 & 62.8 & 46.4 & 54.0 & 52.0 & 58.0 & 52.0 & 47.3 \textcolor{red}{(-4.4)} \\

\rowcolor{cyan!15}
\textbf{Reasmory (GPT-5-mini)} 
& 60.2 & \textbf{41.9} & 73.1 & \textbf{50.2} & \textbf{52.0} & \textbf{55.6} & 42.0 & 71.4 & \textbf{55.8 (+6.2)} \\

\rowcolor{cyan!15}
\textbf{Reasmory (Gemini-3-flash)} 
& \underline{\textbf{69.0}}
& \underline{\textbf{50.4}}
& 68.4
& \underline{\textbf{62.0}}
& \underline{\textbf{71.4}}
& \underline{\textbf{65.0}}
& 62.0
& 72.0 
& \underline{\textbf{65.0 (+13.4)}} \\

\bottomrule
\end{tabular}

\vspace{-3mm}
\end{table}

\noindent \textbf{Video-based Dynamic Scenes.} Table~\ref{tab:eval-vlm4d} presents results on VLM4D-Real~\cite{zhou2025vlm4d}, which evaluates spatial reasoning in dynamic environments with both camera motion and object motion. For this setting, Reasmory uses Flow3r~\cite{cong2026flow3r} to estimate geometry and camera poses across frames, as described in Sec.~\ref{sec:exp_setup}. Compared with static-scene settings, dynamic scenes introduce additional temporal complexity because models must reason jointly about scene geometry and motion. Spatially fine-tuned models~\cite{wu2026spatialmllm,li2026spatialladder,VQASynth} perform poorly in this setting, suggesting that dynamic video reasoning remains difficult for models trained primarily on static or simpler spatial QA data. Among test-time scaling approaches, MindJourney~\cite{yang2026mindjourney} fails to improve performance and often degrades accuracy. Generated videos may help simulate plausible novel viewpoints in static scenes, but they struggle to preserve object dynamics and temporal consistency in videos with camera and object motion. Think3D~\cite{zhang2026think3d} achieves moderate gains over direct GPT-5-mini inference, indicating that explicit 3D reasoning can still help in dynamic environments. However, its gains remain limited, likely because it relies on unconstrained interaction with reconstruction tools. In contrast, Reasmory achieves the best overall performance for both frontier backbones. It improves GPT-5-mini from 65.3\% to 72.7\% (+7.4) and Gemini-3-flash from 76.0\% to 82.0\% (+6.0). These results suggest that constrained program execution over spatial memory remains useful even when camera and object motion introduce additional temporal ambiguity.
\begin{table}[t]
\centering
\scriptsize
\caption{
\textbf{Evaluation results on video-input VLM4D-Real.} We compare methods under each base model separately. 
Bold numbers indicate the best result within the same base model family, and underlined numbers indicate the best result overall.
}
\label{tab:eval-vlm4d}
\begin{tabular}{lccc}
\toprule
\textbf{Method} & \textbf{Ego-centric $\uparrow$}& \textbf{Exo-centric $\uparrow$}& \textbf{Overall $\uparrow$} \\
\midrule
GPT-5-mini & 61.4 & 68.8 & 65.3\\
Gemini-3-flash & 84.0 & 68.0 & 76.0\\
\midrule
\multicolumn{2}{l}{\textit{\textbf{Spatial Models}}}
\\
\midrule
SpaceOM-3B~\cite{VQASynth} & 34.0 & 42.9 & 38.5\\
Spatial-MLLM-4B~\cite{wu2026spatialmllm} & 42.0 & 18.0 & 30.0 \\
SpatialLadder-3B~\cite{li2026spatialladder} & 38.0 & 40.0 & 39.0\\
\midrule
\multicolumn{2}{l}{\textit{\textbf{Test-time Scaling Methods}}}\\
\midrule
Mindjourney (GPT-5-mini)~\cite{yang2026mindjourney} &46.0 & 54.0 & 50.0 \textcolor{red}{\textbf{(-15.3)}} \\
Mindjourney (Gemini-3-flash)~\cite{yang2026mindjourney} & 66.0 & 60.0 & 63.0 \textcolor{red}{(-13.0)}\\
Think3D (GPT-5-mini)~\cite{zhang2026think3d} & 65.3 & 70.0 & 67.6 (+2.3)\\
Think3D (Gemini-3-flash)~\cite{zhang2026think3d} & 72.0 & 66.0 & 69.0 \textcolor{red}{(-7.0)}\\
\rowcolor{cyan!15}
\textbf{Reasmory (GPT-5-mini)} & \textbf{71.4} & \textbf{74.0} & \textbf{72.7 (+7.4)} \\
\rowcolor{cyan!15}
\textbf{Reasmory (Gemini-3-flash)} & \textbf{88.0} & \textbf{76.0 }& \textbf{82.0 (+6.0)}\\
\bottomrule
\end{tabular}
\vspace{-3mm}
\end{table}

\subsection{Ablation Study}
\label{sec:exp_ablation}
\begin{table}[t]
\centering
\scriptsize
\setlength{\tabcolsep}{6pt}
\vspace{2mm}
\caption{Ablations in three settings: vanilla VLMs, VLMs augmented with spatial primitives without verification, and VLMs augmented with spatial primitives and DSL verifier.}
\begin{tabular}{llccc}
\toprule
\textbf{Setting} & \textbf{Model} & \textbf{MindCube} & \textbf{VSI-Bench} & \textbf{VLM4D} \\
\midrule

\multirow{2}{*}{Vanilla VLM}
& GPT    & 58.0 & 49.6 & 65.3 \\
& Gemini & 68.7 & 51.6 & 76.0 \\

\multirow{2}{*}{+ Primitives (no verifier)}
& GPT    & 53.2 & 45.3 & 70.4 \\
& Gemini & 63.5 & 50.4 & 73.0\\

\multirow{2}{*}{+ Primitives + DSL verifier}
& GPT    & \textbf{76.0} & \textbf{55.8} & \textbf{72.7} \\
& Gemini & \underline{\textbf{86.0}} & \underline{\textbf{65.0}} & \underline{\textbf{82.0}} \\
\bottomrule

\end{tabular}
\label{tab:ablation}
\end{table}
Table~\ref{tab:ablation} evaluates the role of DSL verification in Reasmory. We compare direct VLM inference, VLMs augmented with spatial primitives but without DSL verification, and the full Reasmory pipeline with both spatial primitives and DSL verification.

Directly exposing spatial primitives to VLMs without constrained interaction leads to inconsistent results. It degrades GPT-5-mini on MindCube and VSI-Bench, from 58.0\% to 53.2\% and from 49.6\% to 45.3\%, respectively. For Gemini-3-flash, spatial primitives without verification also reduce performance relative to direct inference on all three benchmarks. Although these primitives can help in some cases, such as GPT-5-mini on VLM4D, the gains remain smaller than those of the full pipeline. Adding the DSL verifier consistently gives the best performance across all benchmarks and both frontier models. These results show that Reasmory's gains do not come simply from exposing additional spatial primitives, but from constraining and verifying how models interact with spatial memory.

\subsection{Reliability Analysis}
\label{sec:exp_reliability}

We further analyze the reliability of two key parts of Reasmory: semantic 3D grounding for object-level memory, and planner reliability for generating executable DSL programs.

\noindent \textbf{3D Grounding Reliability.}
To evaluate semantic grounding in spatial memory, we compare our semantic augmentation pipeline with MaskClustering~\cite{yan2024maskclustering}, a representative mask-merging approach. Table~\ref{tab:grounding} reports open-vocabulary 3D grounding results on ScanNet~\cite{dai2017scannet}. We randomly sample 20 scenes and evaluate 10 object categories.
MaskClustering performs reasonably under dense-view settings (231 frames per scene), but degrades substantially under sparse observations. In the sparse-view setting, our method improves mAP50 from 16.7 to 38.4 and mAP25 from 35.7 to 68.4.
We attribute this improvement to two factors. First, SAM3~\cite{carion2025sam} provides stronger open-vocabulary segmentation. For example, MaskClustering fails completely on categories such as \emph{door}, while our method still recovers valid object instances. Second, our geometry-guided cross-view merging is more robust under sparse multi-view observations, matching the practical input setting of VLM reasoning pipelines. These results suggest that our semantic augmentation strategy produces more reliable object-level spatial memory under limited observations.

\noindent \textbf{Planner Reliability.}
Table~\ref{tab:pass-ratio} reports the planner's program validation pass rate on debiased MindCube. We evaluate pass@k, where a sample is considered successful if at least one valid program is generated within $k$ attempts.
Both GPT-5-mini and Gemini-3-flash achieve high pass@1 scores, indicating that frontier VLMs can often generate valid structured spatial reasoning programs. The pass rate further improves with compiler feedback: GPT-5-mini increases from 82.6\% at pass@1 to 95.3\% at pass@3, while Gemini-3-flash reaches 100\% validity within three attempts.
These results show that the DSL verifier and compiler feedback mechanism make planning reliable in practice, enabling most generated programs to become executable after only a few attempts.
\begin{table*}[t]
\centering
\caption{
Reliability analysis for 3D semantic grounding and planner's behavior in Reasmory. Our SAM-3 based instance merging achieves high mAP under sparse input settings. The feedback loop for plan generation guarantees most of the plans are valid after three attempts.
}
\vspace{1mm}
\begin{minipage}[t]{0.61\textwidth}
\centering
\scriptsize

\begin{tabular}{lcccc}
\toprule
 & \textbf{Frames} & \textbf{mAP} & \textbf{mAP50} & \textbf{mAP25} \\
\midrule
MaskClustering (dense)~\cite{yan2024maskclustering} & 231 & \textbf{19.0} & 33.9 & 51.0\\
MaskClustering (sparse)~\cite{yan2024maskclustering} & \textbf{32} & 5.9 & 16.7 & 35.7 \\
\textbf{Ours} & \textbf{32} & \underline{12.1} & \textbf{38.4} & \textbf{68.4} \\
\bottomrule
\label{tab:grounding}
\end{tabular}

{\footnotesize (a) 2D mask merging result on ScanNet subset.}

\end{minipage}
\hfill
\begin{minipage}[t]{0.38\textwidth}
\centering
\scriptsize
\begin{tabular}{lcc}
\toprule
  & \textbf{GPT-5-mini} & \textbf{Gemini-3-flash} \\
\midrule
 \textbf{pass@1} & 82.6 & 78.0 \\
 \textbf{pass@2} & 93.3 & 98.0 \\
 \textbf{pass@3} & 95.3 & 100.0 \\

\bottomrule
\label{tab:pass-ratio}
\end{tabular}

{\footnotesize (b) Planner's pass ratio on MindCube-Tiny debiased version.}
\end{minipage}
\label{tab:reliability}
\end{table*}

\vspace{-5mm}
\subsection{End-to-End Reasmory Reasoning Example}
\label{sec:exp_example}

\begin{figure}[t]
  \centering
   \includegraphics[width=\linewidth]{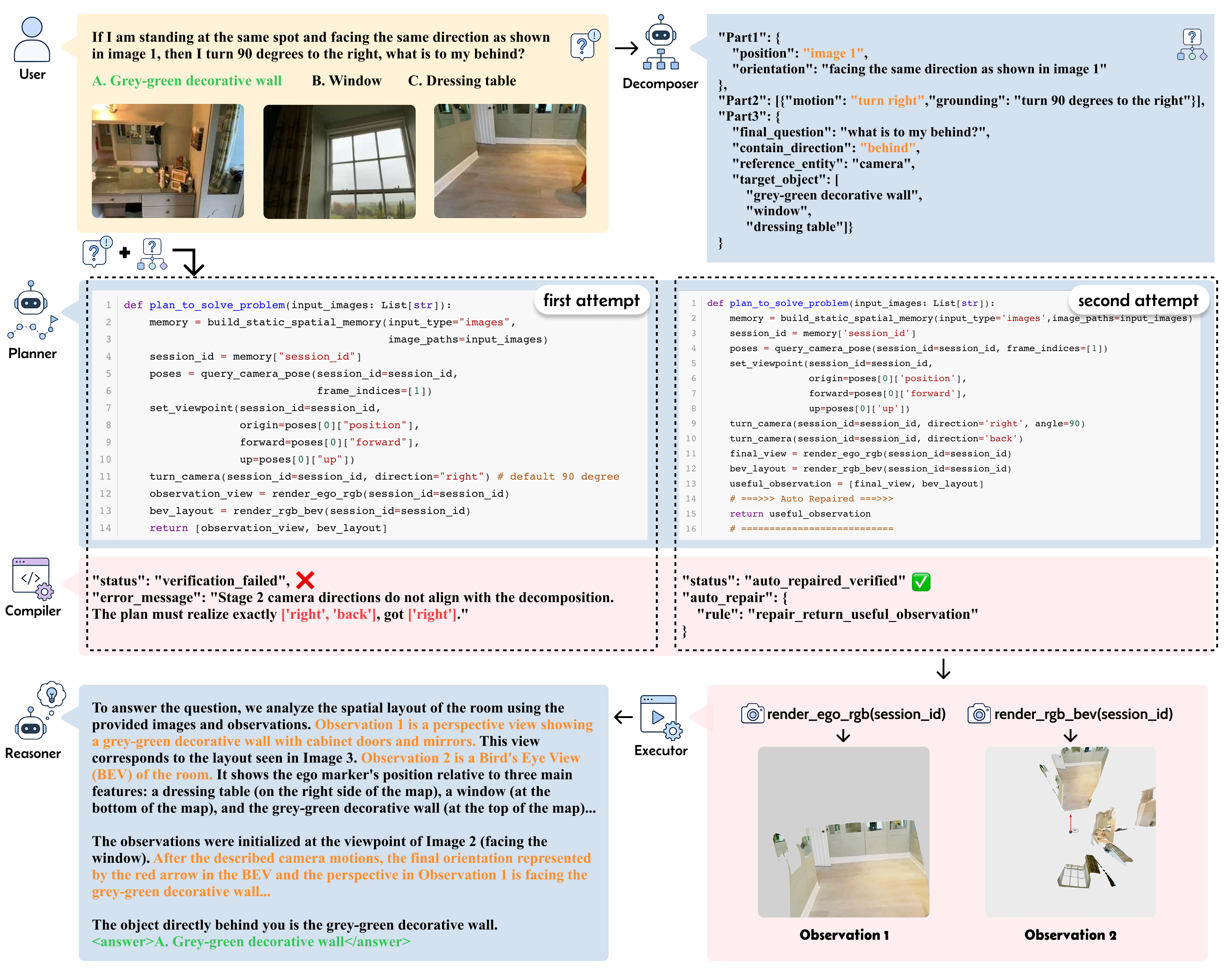}
   \vspace{-5mm}
  \caption{\textbf{An end-to-end reasoning example.}
The trajectory illustrates how verification and repair refine a DSL plan before spatial-memory execution; see Sec.~\ref{sec:exp_example} for details.}
   \label{fig:example}
   \vspace{-5mm}
\end{figure}

Figure~\ref{fig:example} illustrates one complete Reasmory trajectory, from decomposition to verified execution and final answering.
Given a spatial reasoning question, the decomposer first extracts a structured representation of the task, including the reference viewpoint, required viewpoint transformations, and relevant entities. In this example, the decomposition identifies the initial viewpoint (Image~1), the required camera motion (turn right followed by reasoning about what is behind the observer), and the candidate objects involved in the final answer.
Conditioned on both the original question and the decomposition, the planner generates a DSL program for querying the spatial memory. The first generated plan builds the memory and performs a viewpoint transformation, but it omits the directional constraint encoded by \emph{behind}. As a result, the generated camera trajectory does not match the decomposition. Instead of executing an incorrect plan, the compiler detects the mismatch against the decomposition and rejects the program with explicit feedback.
After receiving the error message, the planner generates a revised program that correctly implements the required viewpoint transformation. The revised program still contains a minor implementation error, a missing return statement, which the compiler repairs automatically before validation. After verification, the execution engine deterministically executes the program and renders both egocentric and bird's-eye-view observations corresponding to the transformed viewpoint.
The resulting observations show that the final viewpoint faces the grey-green decorative wall. Based on this enriched context, the reasoner infers the spatial relationship and produces the correct answer. This example highlights two reliability benefits of Reasmory: decomposition-guided verification catches plans that deviate from the intended spatial reasoning, and compiler-assisted correction turns local program errors into recoverable steps before execution.

\vspace{-4mm}
\section{Conclusion}

%


We introduce \textbf{Reasmory}, a framework that uses 3D reconstruction as explicit spatial memory for VLM spatial
reasoning. Reasmory builds memory from multi-view images or videos, augments it with semantically grounded object
instances, and constrains model interaction through validated DSL programs. This reduces brittleness from redundant visual evidence and unconstrained tool use, which can cause invalid or inconsistent spatial operations.
Across multi-view image, static-scene video, and dynamic-scene video benchmarks, Reasmory improves frontier VLMs and
outperforms spatially fine-tuned models and test-time scaling baselines. Ablations show that the gains come not from spatial primitives alone, but from verifying how models query, transform, and render memory. Reliability analyses show effective grounding under sparse observations and high planner validity after compiler feedback.
Reasmory still depends on reconstruction and grounding quality, and may fail with ambiguous grounding, heavy
occlusion, or complex dynamic interactions beyond the recovered memory. Future work may improve dynamic-memory
fidelity, add uncertainty estimates, and extend the DSL for richer spatial and temporal reasoning.


\clearpage
\newpage
\bibliography{egbib}
\end{document}